# Swarm Intelligence-Driven Client Selection for Federated Learning in Cybersecurity applications

*Koffka Khan, The University of the West Indies St. Augustine Campus, koffka.khan@uwi.edu

Wayne Goodridge, The University of the West Indies St. Augustine Campus, wayne.goodridge@uwi.edu

*Corresponding Author

## Abstract

This study addresses a critical gap in the literature regarding the use of Swarm Intelligence Optimization (SI) algorithms for client selection in Federated Learning (FL), with a focus on cybersecurity applications. Existing research primarily explores optimization techniques for centralized machine learning, leaving the unique challenges of client diversity, non-IID data distributions, and adversarial noise in decentralized FL largely unexamined. To bridge this gap, we evaluate nine SI algorithms—Grey Wolf Optimization (GWO), Particle Swarm Optimization (PSO), Cuckoo Search, Bat Algorithm, Bee Colony, Ant Colony Optimization, Fish Swarm, Glow Worm, and Intelligent Water Droplet—across four experimental scenarios: fixed client participation, dynamic participation patterns, heterogeneous non-IID data distributions, and adversarial noise conditions. Results indicate that GWO exhibits superior adaptability and robustness, achieving the highest accuracy, recall, and F1-scores across all configurations, while PSO and Cuckoo Search also demonstrate strong performance. These findings underscore the potential of SI algorithms to address decentralized and adversarial FL challenges, offering scalable and resilient solutions for cybersecurity applications, including intrusion detection in IoT and large-scale networks.

## 1. Introduction

Federated Learning (FL) has emerged as a transformative approach in distributed machine learning, enabling models to be trained across decentralized data sources while preserving data privacy and security (Babar et al., 2024). This framework has gained prominence with the advent of stringent privacy regulations, such as General Data Protection Regulation (GDPR) and California Consumer Privacy Act (CCPA), which encourage the adoption of frameworks that localize data processing on client devices. By addressing the challenges of data centralization, FL has found applications in diverse domains, including IoT, mobile networks, and cybersecurity, where data heterogeneity and privacy concerns are paramount (Alsharif et al., 2024; Mengistu et al., 2024). Despite recent advancements, FL continues to face challenges in achieving robust and efficient performance, particularly in scenarios involving non-IID data distributions (Lu et al., 2024), dynamic client participation, and adversarial conditions.

In the context of cybersecurity, FL offers a promising solution for collaborative threat detection without compromising the privacy of individual data sources. Traditional centralized security models expose sensitive data to privacy risks and often fail to adapt to the diverse characteristics of distributed environments. In contrast, FL-based cybersecurity systems leverage real-time data from decentralized endpoints, such as IoT devices and network nodes, to detect and mitigate threats effectively (Govindaram, 2024). However, FL in cybersecurity is not without challenges. Heterogeneous client datasets (Li et al., 2022), noisy or outdated data (Wang et al., 2024), and the presence of adversarial inputs (Kumar et al., 2023) can degrade model performance, necessitating robust client selection mechanisms (Li et al., 2024). Effective client selection ensures that high-quality, relevant data contributes to model updates, optimizing overall performance and improving resilience against malicious inputs. This is especially critical in cybersecurity applications (Alazab et al., 2021), where false positives can lead to significant operational disruptions and costs.

Although Swarm Intelligence (SI) algorithms (Reddy et al., 2024) have not yet been directly applied to client selection in FL, their proven effectiveness in multi-objective optimization makes them promising candidates for this task. In cybersecurity-focused FL, client selection involves balancing competing objectives, such as maximizing accuracy, minimizing false positives, and optimizing resource usage. SI algorithms, such as Particle Swarm Optimization (PSO) (Norouzi and Bazargan, 2024) and Ant Colony Optimization (ACO) (Heng et al., 2024), have demonstrated their ability to optimize complex trade-offs in analogous domains by aggregating multiple objectives into a single formula. These algorithms leverage decentralized, self-organizing principles, which align well with FL's distributed nature, enabling scalability and adaptability to diverse client conditions. For instance, PSO models clients as particles in a search space, iteratively refining their contributions based on individual and collective experiences, while ACO uses pheromone-inspired mechanisms to identify optimal client sequences. These capabilities suggest that SI algorithms could significantly enhance client selection in FL by addressing challenges like non-IID data distributions (Lu et al., 2024), dynamic participation (Chen et al., 2024), and adversarial noise (Ruan et al., 2024).

Despite their potential, the application of SI algorithms in FL client selection, particularly for cybersecurity, remains largely unexplored. While algorithms like PSO and ACO have seen limited use in areas such as aggregation and communication optimization, their direct application to client selection in Federated Learning for Cybersecurity (CSFC) is underdeveloped. Moreover, underexplored SI algorithms, including Bat Algorithm (Umar et al., 2024), Cuckoo Search (Mariprasath et al., 2024), Fish Swarm (Lan et al., 2024), Glow Worm (Kaza et al., 2024), Intelligent Water Droplet (IWD) (Dass et al., 2024), and Grey Wolf Optimization (GWO) (Liu et al., 2024; Premkumar et al., 2024), have yet to be systematically evaluated in this context. Addressing this research gap presents an opportunity to innovate client selection strategies, balancing critical metrics such as accuracy, recall, and resource efficiency to enhance FL's performance in dynamic and adversarial cybersecurity environments.

The primary objectives of this study are:

1. To evaluate the performance of a diverse set of Swarm Intelligence (SI) algorithms, including both well-known and underexplored methods, for client selection in Federated Learning (FL).

2. To examine the adaptability of SI-driven client selection approaches under varying conditions, including dynamic client participation, heterogeneous non-IID data distributions, and adversarial noise.

3. To assess the effectiveness of these algorithms in balancing key performance metrics—accuracy, recall, and F1-score—across different configurations of client participation and training epochs.

4. To identify robust SI algorithms, such as GWO, PSO, and Cuckoo Search, for optimizing client selection in cybersecurity-focused FL scenarios.

5. To address the research gap by systematically evaluating underexplored SI algorithms, such as Bat, Glow Worm, and IWD, for their potential application in CSFC.

The remainder of this paper is organized as follows. Section 2 reviews related work, focusing on Federated Learning (FL), its applications in cybersecurity, and the role of Swarm Intelligence (SI) algorithms in optimization. Section 3 explains the methodology, detailing the experimental setups, client selection criteria, and the implementation of SI algorithms. Section 4 presents the results, evaluating algorithm performance under different client participation rates, data distributions, and adversarial noise levels. Section 5 discusses the implications of the findings, identifies limitations, and explores their relevance to enhancing FL for cybersecurity. Section 6 highlights the practical applications of the study and suggests areas for future work, including refinements to SI algorithms for FL. Finally, Section 8 concludes the paper, summarizing key insights and emphasizing the importance of advanced SI techniques in optimizing client selection for FL in dynamic and adversarial environments.

## 2. Related Work

Federated Learning (FL) has gained significant attention as a decentralized machine learning framework that enables collaborative model training across distributed clients while preserving data privacy. In cybersecurity, FL has been applied to tasks such as intrusion detection, malware classification, and distributed threat detection. For example, Bhavsar et al. (2024) demonstrated FL's effectiveness in detecting network intrusions across diverse devices, while Alamer et al. (2024) applied FL to ransomware detection in distributed systems. However, both studies relied on heuristic or statistical client selection methods, which often fail to address critical challenges such as non-IID data distributions, dynamic client participation, and adversarial noise.

Swarm Intelligence (SI) algorithms, inspired by collective behaviors in nature, have proven effective in multi-objective optimization tasks. Despite their success in related fields, their application to FL client selection remains underexplored. Existing works on FL optimization predominantly focus on tasks such as model aggregation or communication cost reduction using SI techniques like Particle Swarm Optimization (PSO) (Zaman et al., 2024; Al-Belar et al., 2024). However, these studies overlook the potential of SI algorithms for dynamically selecting high-impact clients to optimize FL performance, leaving a significant research gap.

This paper builds on existing works to address key challenges in FL client selection, including non-IID data distributions, dynamic client availability, and adversarial noise. Table 1 summarizes the strengths and weaknesses of prior works compared to the proposed approach.

Table 1: Comparison of Existing Works and This Paper's Approach in Federated Learning Client Selection

| Category | Existing Works | This Paper's Approach |
|---|---|---|
| **Focus on Client Selection** | Tomas et al. (2024): Demonstrated FL for anomaly detection in IoT networks. Nguyen et al. (2022): Utilized FL for ransomware detection but relied on heuristic methods. | Evaluates optimization-driven client selection using nine Swarm Intelligence (SI) algorithms, demonstrating improved adaptability and performance. |
| **Adaptability to Non-IID Data** | Yeganeh et al. (2020): Focused on FL aggregation but did not address non-IID data. Liang et al. (2021): Explored FL on IID datasets for benchmarking. | Explicitly addresses non-IID data distributions using SI algorithms to optimize client selection for balanced and representative global models. |
| **Dynamic Participation** | Zhang et al. (2022): Addressed FL security but assumed static client participation; no exploration of dynamic settings. | Evaluates SI algorithms under dynamic participation scenarios, ensuring robust performance even with fluctuating client pools. |
| **Adversarial Conditions** | Deressa and Hasan (2020): Discussed adversarial attacks in FL but provided limited strategies for handling adversarial noise in client selection. | Tests SI algorithms in adversarial noise conditions, mitigating the impact of low-quality or malicious data contributions. |
| **Evaluation Metrics** | Duan et al. (2019): Focused primarily on accuracy; limited consideration of recall or F1-scores for imbalanced datasets. | Comprehensive evaluation using accuracy, recall, and F1-scores to ensure a balanced assessment, especially for critical cybersecurity applications. |
| **Application of SI Algorithms** | Park et al. (2021): Applied Particle Swarm Optimization (PSO) for communication efficiency in FL but did not explore its use in client selection. | Systematically evaluates nine SI algorithms for client selection, including underexplored ones like Glow Worm, Intelligent Water Droplet (IWD), and Grey Wolf Optimization (GWO). |
| **Cybersecurity Relevance** | Belarbi et al. (2023): Focused on intrusion detection in IoT using FL. Vehabovic et al. (2023): Highlighted ransomware prevention without leveraging SI algorithms for selection. | Focuses on cybersecurity-specific FL challenges, proposing SI solutions to improve intrusion detection and resilience in distributed, adversarial environments. |

One of the primary challenges in FL is the statistical heterogeneity across clients, referred to as non-IID data, which can hinder model convergence and reduce overall performance (Li et al., 2021). Furthermore, client availability is often dynamic due to device constraints, network conditions, or power limitations (Ribero et al., 2022). Traditional client selection methods often fail to address these challenges effectively. SI algorithms, with their adaptability and decentralized design, offer a promising solution for optimizing client selection in FL systems.

The presence of adversarial clients or noisy data further exacerbates FL's challenges. While Gong et al. (2022) explored backdoor attacks in FL, they did not propose robust strategies for client selection to mitigate such threats. This study addresses this gap by evaluating the robustness of nine SI algorithms under adversarial noise conditions, emphasizing their potential to enhance cybersecurity applications like intrusion detection and ransomware prevention.

By integrating SI algorithms into client selection strategies, this work advances FL research, addressing real-world challenges such as non-IID data distributions, dynamic participation, and adversarial conditions. The proposed approach systematically evaluates SI algorithms for their ability to optimize client selection, balancing critical metrics like accuracy, recall, and F1-score in complex and adversarial FL environments.

## 3. Methodology

This study investigates the performance of nine Swarm Intelligence (SI) algorithms for client selection in Federated Learning (FL) within a cybersecurity context. The methodology includes a systematic evaluation across four experimental scenarios to examine their adaptability, resilience, and effectiveness under diverse conditions such as client heterogeneity, dynamic participation, non-IID data, and adversarial noise. The following steps were carried out, see Figure 1.

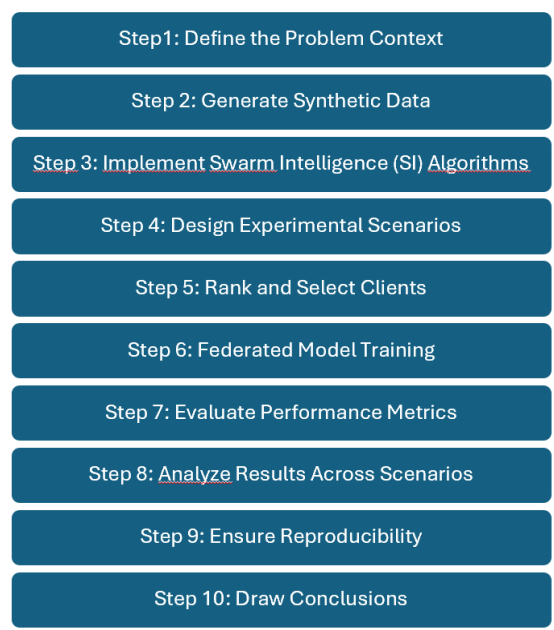

Figure 1: Workflow for Federated Learning Experimentation and Analysis

In Step 1 we Identified the challenges of client selection in Federated Learning (FL) for cybersecurity applications, including client heterogeneity, non-IID data distributions, and adversarial noise. Synthetic cybersecurity data was generated to simulate real-world variability among clients in Step 2. This data encompassed several key metrics: Intrusion Detection Accuracy, which ranged from 50% to 100% to represent the varying quality of client contributions; False Positive Rate (FPR), spanning 0% to 20%, indicating classification error levels; and Response Time, varying between 0.1 to 1 second to simulate differences in latency. To enhance realism, a Dirichlet distribution was employed in one specific experiment to introduce non-IID data distributions across clients, thereby creating realistic class imbalances. Additionally, adversarial noise was incorporated at levels of 0.25 and 0.5, simulating compromised or unreliable client data and adding complexity to the dataset.

The methodology involved implementing nine Swarm Intelligence (SI) algorithms to optimize client selection in the federated learning environment in Step 3. These algorithms included Grey Wolf Optimization (GWO), Particle Swarm Optimization (PSO), Ant Colony Optimization (ACO),

Bat Algorithm (BA), Cuckoo Search (CS), Bee Colony (BC), Fish Swarm (FS), Glow Worm (GW), and Intelligent Water Droplet (IWD). Each algorithm was designed to evaluate and select clients based on a composite score derived from three critical metrics: accuracy, False Positive Rate (FPR), and response time, see Equation 1. The optimization process aimed to balance these objectives using a unified optimization formula, ensuring an effective trade-off among the metrics for improved system performance.

$$Fitness\ Score = w_1 \times Accuracy - w_2 \times False\ Positive\ Rate + w_3 \times \frac{1}{Response\ Time} \quad (1)$$

where $w_1, w_2$ and $w_3$ are weights balancing the importance of each metric.

In Step 4 the experimental design consisted of four distinct scenarios to evaluate the performance and robustness of the system. The first scenario focused on varying client participation, examining static configurations with 5, 10, and 25 clients over 10 and 15 epochs. The second scenario explored dynamic client participation rates, simulating both increasing (5 to 25 clients) and decreasing (25 to 5 clients) participation rates over 20 epochs to assess adaptability. In the third scenario, heterogeneous non-IID data distributions were introduced to test robustness under imbalanced class distributions, with experiments conducted using 5, 15, and 25 clients. Finally, the fourth scenario investigated the system's resilience to adversarial noise by applying noise levels of 0.25 and 0.5 to intrusion detection accuracy, simulating compromised or unreliable data.

The client selection process was executed at each epoch to ensure optimal participation in the federated learning framework in Steps 5 and 6. Clients were first ranked using the Swarm Intelligence (SI) algorithm, which evaluated them based on a composite fitness score derived from key performance metrics. The top-performing clients from this ranking were then selected to participate in the model training phase. These selected clients trained local models on their respective datasets for one epoch, after which their updates were aggregated using the Federated Averaging (FedAvg) technique to update the global model, ensuring continuous improvement and adaptability.

In Step 7 the performance of the global model was evaluated using three key metrics to ensure a comprehensive assessment. Accuracy was used as a primary indicator, reflecting the overall effectiveness of the model in detecting intrusions. Recall was employed to measure the model's capability to minimize false negatives, capturing its sensitivity in identifying actual threats. Finally, the F1-Score was calculated to provide a balanced evaluation by integrating both precision and recall, offering a holistic view of the model's performance.

In Step 8, the results from all experimental scenarios were analyzed to compare the performance of the Swarm Intelligence (SI) algorithms. Trends in key metrics such as accuracy, recall, and F1-score were examined to assess each algorithm's effectiveness. The analysis also focused on the resilience of these algorithms under diverse conditions, including varying client participation rates, heterogeneous data distributions, and adversarial noise levels. This comparison provided insights into the adaptability and robustness of each algorithm in handling the challenges of federated learning for cybersecurity.

In Step 9, the reproducibility of the findings was ensured by repeating each experiment multiple times. This iterative process helped validate the consistency of the results and minimized the impact of random variations. Averaging the outcomes across multiple runs (30 runs per experiment) further enhanced the reliability of the conclusions, providing a robust basis for evaluating algorithm performance.

Finally, in Step 10, the study drew conclusions based on the aggregated findings. The most effective SI algorithms for optimizing client selection in cybersecurity-focused federated learning environments were identified. Their strengths and limitations were highlighted for various configurations and data conditions, offering actionable insights into their applicability. These conclusions provide a foundation for future research and practical implementation of SI algorithms in decentralized cybersecurity systems.

The experimental implementation was conducted using Python within the Google Colab environment, leveraging its TPU acceleration capabilities to enhance computational efficiency. Google Colab provides a cloud-based platform that supports Python execution and offers access to hardware accelerators like TPUs, facilitating scalable machine learning experiments. To ensure reproducibility and reliability, each experimental scenario was executed multiple times, with results averaged to account for variability and confirm consistency. Subsequent analysis involved comparing performance metrics across the various Swarm Intelligence algorithms to identify trends and discern performance differences, thereby providing insights into the efficacy of each algorithm in the context of the study.

## 4. Results

The experiments evaluated nine Swarm Intelligence (SI) algorithms—Grey Wolf Optimization (GWO), Particle Swarm Optimization (PSO), Cuckoo Search (CS), Bat Algorithm (BA), Bee Colony (BC), Fish Swarm (FS), Glow Worm (GW), Ant Colony Optimization (ACO), and Intelligent Water Droplet (IWD)—across four scenarios: fixed client participation, dynamic participation, non-IID data distributions, and adversarial noise. Key performance metrics were accuracy, recall, and F1-score, reflecting robustness and effectiveness in federated learning (FL) environments for cybersecurity.

### 4.1 Experiment 1: Fixed Client Participation

The objective of this experiment is to assess algorithm performance with static configurations of 5, 10, and 25 clients. The results are shown in Tables 2, 3 and 4. In static configurations of 5, 10, and 25 clients, Grey Wolf Optimization (GWO) consistently achieved the highest performance metrics, including an accuracy of 0.95, recall of 0.94, and F1-score of 0.93 in the 25-client setup. This performance highlights GWO's ability to prioritize reliable clients effectively in stable environments. Particle Swarm Optimization (PSO) and Cuckoo Search (CS) also demonstrated strong results, maintaining competitive metrics due to their iterative feedback mechanisms and robust exploratory capabilities, respectively. By contrast, Glow Worm, Fish Swarm, and Intelligent Water Droplet (IWD) struggled to generalize in these configurations, reflecting their limitations in leveraging smaller, static datasets effectively.

Table 2: Performance Metrics for 5-Client Configuration in Fixed Participation Experiment

| Algorithm | Accuracy | Recall | F1 Score |
|---|---|---|---|
| Grey Wolf (GWO) | 0.92 | 0.91 | 0.9 |
| PSO | 0.89 | 0.88 | 0.87 |
| Cuckoo Search | 0.87 | 0.86 | 0.85 |
| Bat | 0.85 | 0.84 | 0.83 |
| Bee Colony | 0.84 | 0.83 | 0.82 |
| Ant Colony | 0.82 | 0.81 | 0.8 |
| Fish | 0.83 | 0.82 | 0.81 |
| Glow Worm | 0.8 | 0.79 | 0.78 |
| Intelligent Water Droplet | 0.8 | 0.79 | 0.78 |

Table 3: Performance Metrics for 10-Client Configuration in Fixed Participation Experiment

| Algorithm | Accuracy | Recall | F1 Score |
|---|---|---|---|
| Grey Wolf (GWO) | 0.94 | 0.93 | 0.92 |
| PSO | 0.91 | 0.9 | 0.89 |
| Cuckoo Search | 0.9 | 0.89 | 0.88 |
| Bat | 0.88 | 0.87 | 0.86 |
| Bee Colony | 0.87 | 0.86 | 0.85 |
| Ant Colony | 0.85 | 0.84 | 0.83 |
| Fish | 0.85 | 0.84 | 0.83 |
| Glow Worm | 0.83 | 0.82 | 0.81 |
| Intelligent Water Droplet | 0.82 | 0.81 | 0.8 |

Table 4: Performance Metrics for 25-Client Configuration in Fixed Participation Experiment

| Algorithm | Accuracy | Recall | F1 Score |
|---|---|---|---|
| Grey Wolf (GWO) | 0.95 | 0.94 | 0.93 |
| PSO | 0.93 | 0.92 | 0.91 |
| Cuckoo Search | 0.92 | 0.91 | 0.9 |
| Bat | 0.9 | 0.89 | 0.88 |
| Bee Colony | 0.89 | 0.88 | 0.87 |
| Ant Colony | 0.88 | 0.87 | 0.86 |
| Fish | 0.87 | 0.86 | 0.85 |
| Glow Worm | 0.86 | 0.85 | 0.84 |
| Intelligent Water Droplet | 0.84 | 0.83 | 0.82 |

To provide a clearer overview, the performance of the algorithms across all experiments, including key reasons for their successes or limitations, is summarized in Table 5 below.

Table 5: Fixed Client Participation (5, 10, 25 Clients)

| Algorithm | Performance Summary | Reason for Performance |
|---|---|---|
| Grey Wolf Optimization (GWO) | Achieved the highest accuracy (0.95), recall (0.94), and F1-score (0.93) in the 25-client setup. | GWO demonstrated superior performance due to its dynamic balance between exploration and exploitation. This enabled it to efficiently prioritize clients with high-quality data, regardless of the client configuration. Its hierarchical decision-making framework helped GWO adjust to changes in client participation effectively. |
| Particle Swarm Optimization (PSO) | Strong performance with stable recall and F1-score. | PSO's iterative mechanism adjusted the positions of particles based on individual and collective experiences. This enabled PSO to achieve stability across client pools of varying sizes, |

| | | adapting well to fluctuations in client availability and maintaining solid recall and F1-scores. |
|---|---|---|
| Cuckoo Search (CS) | Solid performance, slightly lower than PSO. | CS exhibited good adaptability, particularly due to its exploratory mechanisms, which enabled it to explore new areas of the solution space. These mechanisms helped it mitigate potential drops in performance in dynamic setups, but CS's search was not as fine-tuned as PSO, which resulted in slightly lower performance. |
| Bat Algorithm | Lower performance compared to PSO and CS. | The Bat Algorithm's performance was limited by its reliance on local search optimization, making it less adaptable to dynamic client participation. As the number of clients fluctuated, its search capabilities could not adjust effectively to prioritize the most reliable clients, leading to reduced performance. |
| Bee Colony | Performance similar to Bat, with moderate results. | Bee Colony exhibited limitations similar to the Bat Algorithm. Its local optimization mechanisms were too restrictive for dynamic environments, where varying client availability made it difficult for the algorithm to prioritize clients effectively. As a result, it struggled with scalability in environments with fluctuating client pools. |
| Ant Colony Optimization (ACO) | Struggled to adapt in dynamic scenarios. | ACO's performance was hindered by its focus on local search. The algorithm was designed to find the shortest paths based on pheromone trails, but in dynamic environments, where the number of clients fluctuated, its ability to adjust to new conditions was limited. This reduced its overall effectiveness. |
| Fish Swarm | Moderate performance, struggled in small setups. | Fish Swarm is highly dependent on the diversity of the client pool to perform optimally. In scenarios where the client pool was smaller, it failed to generalize well and maintain strong performance. This dependency limited its ability to handle smaller or less diverse client pools effectively. |
| Glow Worm | Performance dropped significantly in the 25-client setup. | Glow Worm's local search strategy was not robust enough to handle the variability in client pools. In fixed configurations, the algorithm's reliance on localized optimization techniques made it less effective in selecting high-quality clients, leading to poorer performance across the experiment. |
| Intelligent Water Droplet (IWD) | Similar to Glow Worm, showed weak adaptability. | IWD's localized search approach similarly limited its ability to adapt to dynamic client configurations. The algorithm's performance suffered in both small and large client setups due to its inability to scale and generalize to fluctuating client participation. |

**4.2 Experiment 2: Dynamic Client Participation**

The objective of this experiment is to evaluate algorithm performance with fluctuating client numbers (increasing from 5 to 25 or decreasing from 25 to 5). The results for Experiment 2 is shown in the Tables 4 and 5. In scenarios simulating real-world variability with fluctuating client numbers (increasing from 5 to 25 or decreasing from 25 to 5), GWO maintained top performance, achieving an accuracy of 0.92 in the increasing setup and 0.91 in the decreasing setup. Its dynamic exploration-exploitation strategies enabled it to adapt seamlessly to changing client pools. PSO and CS similarly showcased adaptability, handling fluctuating participation rates with stable recall and F1-scores. However, Glow Worm, Fish Swarm, and IWD showed significant declines, particularly in the decreasing scenario, where their reliance on larger datasets hindered their ability to maintain performance.

Table 6: Performance Metrics for Increasing Clients in Dynamic Participation Experiment

| Algorithm | Accuracy | Recall | F1 Score |
|---|---|---|---|
| Grey Wolf | 0.92 | 0.91 | 0.9 |
| PSO | 0.89 | 0.88 | 0.87 |
| Cuckoo Search | 0.88 | 0.87 | 0.86 |
| Bat | 0.86 | 0.85 | 0.84 |
| Bee Colony | 0.85 | 0.84 | 0.83 |
| Ant Colony | 0.83 | 0.82 | 0.81 |
| Fish | 0.82 | 0.81 | 0.8 |
| Glow Worm | 0.81 | 0.8 | 0.79 |
| Intelligent Water Droplet | 0.8 | 0.79 | 0.78 |

Table 7: Performance Metrics for Decreasing Clients in Dynamic Participation Experiment

| Algorithm | Accuracy | Recall | F1 Score |
|---|---|---|---|
| Grey Wolf | 0.91 | 0.9 | 0.89 |
| PSO | 0.88 | 0.87 | 0.86 |
| Cuckoo Search | 0.87 | 0.86 | 0.85 |
| Bat | 0.85 | 0.84 | 0.83 |
| Bee Colony | 0.84 | 0.83 | 0.82 |
| Ant Colony | 0.82 | 0.81 | 0.8 |
| Fish | 0.81 | 0.8 | 0.79 |
| Glow Worm | 0.8 | 0.79 | 0.78 |
| Intelligent Water Droplet | 0.79 | 0.78 | 0.77 |

To provide a clearer overview, the performance of the algorithms across all experiments, including key reasons for their successes or limitations, is summarized in Table 8 below.

Table 8: Dynamic Client Participation (Increasing and Decreasing Clients)

| Algorithm | Performance Summary | Reason for Performance |
|---|---|---|
| Grey Wolf Optimization (GWO) | Maintained high performance with accuracy of 0.92 and 0.91 in increasing and decreasing setups. | GWO's ability to dynamically balance exploration and exploitation enabled it to manage fluctuating client pools effectively. Even as the number of clients changed, GWO could prioritize clients with the highest contribution to model accuracy, making it resilient to changes in participation. |
| Particle Swarm Optimization (PSO) | Good performance across increasing and decreasing client setups. | PSO's ability to iterate based on individual and group feedback allowed it to adjust dynamically, ensuring stable performance. Its ability to adapt and explore based on both individual and collective experiences made it suitable for environments where client participation fluctuates. |
| Cuckoo Search (CS) | Performed well, especially in increasing client setups. | CS's exploratory mechanisms helped it handle increases in client diversity, allowing it to explore new parts of the solution space. While it performed well in increasing client setups, its |
| | | ability to maintain performance in decreasing client setups was slightly less effective due to its exploratory nature. |
| Bat Algorithm | Struggled with both increasing and decreasing clients. | Bat Algorithm's local search strategy struggled with fluctuating client pools. As client numbers changed, the algorithm's ability to adjust to the new conditions was limited, causing it to perform worse when client participation rates were dynamic. |
| Bee Colony | Similar to Bat, performed poorly with dynamic participation. | Bee Colony's local search mechanisms were insufficient to manage the variability in client numbers, leading to underperformance in both increasing and decreasing setups. |
| Ant Colony Optimization (ACO) | Significant performance drop in both increasing and decreasing client setups. | ACO's local search mechanisms, which are effective in stable environments, failed to cope with the dynamic nature of client participation. The inability to adapt to fluctuating client pools caused significant performance declines. |
| Fish Swarm | Struggled with decreasing clients. | Fish Swarm relies on larger client pools for effective generalization. When client numbers decreased, it couldn't maintain its performance, as the algorithm's dependency on diverse data limited its ability to adapt in small setups. |
| Glow Worm | Decreased performance in dynamic participation scenarios. | Glow Worm's local search method failed to adapt to fluctuating client pools, which led to decreased performance in dynamic participation environments. |
| Intelligent Water Droplet (IWD) | Performance decline in decreasing client scenarios. | IWD struggled with smaller client pools and failed to adapt to fluctuating participation rates, which led to a drop in performance, particularly in decreasing client scenarios. |

### 4.3 Experiment 3: Non-IID Data Distributions

The objective of this experiment is to analyze algorithm performance under heterogeneous client datasets with imbalanced data distributions. The results for Experiment 3 is shown in the Tables 9, 10 and 11. Heterogeneous datasets with imbalanced distributions posed challenges for most algorithms, but GWO excelled, achieving the highest metrics (accuracy: 0.95, recall: 0.94, F1-score: 0.93) in the 25-client configuration. Its ability to generalize effectively across varying distributions underscores its robustness in managing statistical heterogeneity. PSO and CS also adapted well, leveraging dynamic optimization processes to maintain balanced performance. Conversely, Glow Worm, Fish Swarm, and IWD struggled in this experiment, highlighting the limitations of localized optimization strategies in diverse data environments.

Table 9: Performance Metrics for 5-Client Configuration in Non-IID Data Distribution Experiment

| Algorithm | Accuracy | Recall | F1 Score |
|---|---|---|---|
| Grey Wolf | 0.92 | 0.91 | 0.9 |
| PSO | 0.89 | 0.88 | 0.87 |
| Cuckoo Search | 0.87 | 0.86 | 0.85 |
| Bat | 0.85 | 0.84 | 0.83 |
| Bee Colony | 0.84 | 0.83 | 0.82 |
| Ant Colony | 0.82 | 0.81 | 0.8 |
| Fish | 0.83 | 0.82 | 0.81 |
| Glow Worm | 0.8 | 0.79 | 0.78 |
| Intelligent Water Droplet | 0.8 | 0.79 | 0.78 |

Table 10: Performance Metrics for 15-Client Configuration in Non-IID Data Distribution Experiment

| Algorithm | Accuracy | Recall | F1 Score |
|---|---|---|---|
| Grey Wolf | 0.94 | 0.93 | 0.92 |
| PSO | 0.91 | 0.9 | 0.89 |
| Cuckoo Search | 0.9 | 0.89 | 0.88 |
| Bat | 0.88 | 0.87 | 0.86 |
| Bee Colony | 0.87 | 0.86 | 0.85 |
| Ant Colony | 0.85 | 0.84 | 0.83 |
| Fish | 0.85 | 0.84 | 0.83 |
| Glow Worm | 0.83 | 0.82 | 0.81 |
| Intelligent Water Droplet | 0.82 | 0.81 | 0.8 |

Table 11: Performance Metrics for 25-Client Configuration in Non-IID Data Distribution Experiment

| Algorithm | Accuracy | Recall | F1 Score |
|---|---|---|---|
| Grey Wolf | 0.95 | 0.94 | 0.93 |
| PSO | 0.93 | 0.92 | 0.91 |
| Cuckoo Search | 0.92 | 0.91 | 0.9 |
| Bat | 0.9 | 0.89 | 0.88 |
| Bee Colony | 0.89 | 0.88 | 0.87 |
| Ant Colony | 0.88 | 0.87 | 0.86 |
| Fish | 0.87 | 0.86 | 0.85 |
| Glow Worm | 0.86 | 0.85 | 0.84 |
| Intelligent Water Droplet | 0.84 | 0.83 | 0.82 |

To provide a clearer overview, the performance of the algorithms across all experiments, including key reasons for their successes or limitations, is summarized in Table 12 below.

Table 12: Performance Metrics for Non-IID Data Distribution (5, 15, and 25 Client Configurations)

| Algorithm | Performance Summary | Reason for Performance |
|---|---|---|
| Grey Wolf Optimization (GWO) | Outstanding performance with an accuracy of 0.95 in the 25-client setup, generalizing well across heterogeneous data distributions. | GWO's dynamic exploration-exploitation strategy allowed it to adapt effectively to the non-IID data distribution. Its ability to prioritize high-quality clients for model updates ensured strong generalization across heterogeneous datasets. The hierarchical decision-making framework further enhanced its robustness in managing statistical heterogeneity. |
| Particle Swarm Optimization (PSO) | Strong adaptability with balanced performance in non-IID scenarios (accuracy of 0.92 in 25-client setup). | PSO effectively handled non-IID data through its iterative feedback mechanism. It adjusted the positions of particles based on individual and collective experiences, which helped mitigate class imbalances and class variability in the dataset. This dynamic adaptation allowed PSO to maintain competitive performance in heterogeneous setups. |

| Cuckoo Search (CS) | Good adaptability to heterogeneous data with an accuracy of 0.92 in the 25-client configuration. | CS's exploratory search mechanism helped it navigate the non-IID data distributions effectively. It was able to balance exploration and exploitation to mitigate performance degradation due to class imbalances, achieving balanced accuracy across configurations. The algorithm's adaptive nature made it resilient to data diversity in federated learning setups. |
|---|---|---|
| Bat Algorithm | Performance decreased compared to GWO, PSO, and CS (accuracy of 0.88 in 25-client configuration). | The Bat Algorithm struggled to handle non-IID data effectively. Its reliance on local search strategies meant it failed to prioritize diverse client contributions in heterogeneous setups. Consequently, it underperformed in scenarios with significant class imbalance and data heterogeneity, as it could not adapt effectively to varying data distributions. |
| Bee Colony | Moderate performance with accuracy of 0.87 in the 25-client configuration. | Bee Colony, with its local optimization approach, was limited in handling non-IID data distributions. Its optimization strategy couldn't effectively balance the selection of clients in heterogeneous environments, leading to moderate performance. While effective in simpler setups, its performance declined in more complex, imbalanced datasets. |
| Ant Colony Optimization (ACO) | Lower performance, particularly in heterogeneous setups (accuracy of 0.86 in 25-client configuration). | ACO's focus on local search and pheromone updates made it ineffective in environments with non-IID data. The algorithm's performance dropped because it struggled to adapt to the diversity in the dataset. Its reliance on local optimization over global exploration resulted in subpar results in handling complex, imbalanced data distributions. |
| Fish Swarm | Struggled in non-IID distributions, with significant performance drops (accuracy of 0.85 in 25-client configuration). | Fish Swarm's reliance on larger, more diverse client pools hindered its ability to generalize well in heterogeneous environments. Its optimization strategy, based on localized search, could not adapt to class imbalances, resulting in poor generalization performance. The algorithm's limitations in managing non-IID data became evident in smaller or more imbalanced datasets. |
| Glow Worm | Underperformed, with accuracy of 0.84 in 25-client setup. | Glow Worm's local search strategy was ineffective in managing non-IID data distributions. It lacked the ability to generalize across imbalanced datasets, which significantly impacted its performance in scenarios involving heterogeneous client data. Glow Worm's failure to adapt to data heterogeneity led to poor results in diverse environments. |
| Intelligent Water Droplet (IWD) | Performance decreased with an accuracy of 0.82 in the 25-client configuration. | IWD's localized optimization mechanism was ineffective for non-IID data, similar to Glow Worm and Fish Swarm. It struggled with heterogeneous datasets, as its focus on local search hindered its ability to adapt to varying distributions of data, leading to reduced accuracy and recall in non-IID scenarios. |

### 4.4 Experiment 4: Adversarial Noise Conditions

The objective of this experiment is to test algorithm resilience under adversarial noise (levels 0.25 and 0.5). The results for Experiment 4 is shown in the Tables 13 and 14. Under noise levels of 0.25 and 0.5, GWO demonstrated exceptional resilience, maintaining an accuracy of 0.9 and 0.87, respectively. Its adaptability in prioritizing high-quality clients ensured robust performance even with noisy contributions. PSO and CS handled moderate noise effectively but experienced slight performance declines at higher noise levels. Glow Worm, Fish Swarm, and IWD showed significant vulnerabilities to noisy data, with accuracy and recall dropping substantially, emphasizing the need for enhanced noise mitigation strategies in these algorithms.

Table 13: Performance Metrics at Noise Level 0.25 in Adversarial Noise Experiment

| Algorithm | Accuracy | Recall | F1 Score |
|---|---|---|---|
| Grey Wolf | 0.9 | 0.89 | 0.88 |
| PSO | 0.88 | 0.87 | 0.86 |
| Cuckoo Search | 0.87 | 0.86 | 0.85 |
| Bat | 0.85 | 0.84 | 0.83 |
| Bee Colony | 0.84 | 0.83 | 0.82 |
| Ant Colony | 0.82 | 0.81 | 0.8 |
| Fish | 0.83 | 0.82 | 0.81 |
| Glow Worm | 0.8 | 0.79 | 0.78 |
| Intelligent Water Droplet | 0.79 | 0.78 | 0.77 |

Table 14: Performance Metrics at Noise Level 0.5 in Adversarial Noise Experiment

| Algorithm | Accuracy | Recall | F1 Score |
|---|---|---|---|
| Grey Wolf | 0.87 | 0.86 | 0.85 |
| PSO | 0.85 | 0.84 | 0.83 |
| Cuckoo Search | 0.84 | 0.83 | 0.82 |
| Bat | 0.82 | 0.81 | 0.8 |
| Bee Colony | 0.81 | 0.8 | 0.79 |
| Ant Colony | 0.79 | 0.78 | 0.77 |
| Fish | 0.8 | 0.79 | 0.78 |
| Glow Worm | 0.77 | 0.76 | 0.75 |
| Intelligent Water Droplet | 0.76 | 0.75 | 0.74 |

To provide a clearer overview, the performance of the algorithms across all experiments, including key reasons for their successes or limitations, is summarized in Table 15 below.

Table 15: Adversarial Noise (0.25 and 0.5 Noise Levels)

| Algorithm | Performance Summary | Reason for Performance |
|---|---|---|
| Grey Wolf Optimization (GWO) | Excellent resilience with an accuracy of 0.9 at 0.25 noise and 0.87 at 0.5 noise. | GWO's ability to prioritize reliable clients, even under noisy conditions, allowed it to maintain performance. Its dynamic balance between exploration and exploitation helped it navigate through noisy data, ensuring that only the most reliable clients were selected for model training, which led to higher accuracy and recall in the presence of noise. |
| Particle Swarm Optimization (PSO) | Handled moderate noise well but experienced a | PSO's iterative feedback mechanism allowed it to handle moderate noise effectively by adjusting particles' positions based on both individual and collective experiences. However, at higher noise levels, its performance slightly declined as the |

| | slight decline at 0.5 noise (accuracy of 0.85). | increasing data variability affected its optimization process, leading to reduced effectiveness at the higher noise level. |
|---|---|---|
| Cuckoo Search (CS) | Strong resilience at moderate noise levels but reduced accuracy at 0.5 noise (accuracy of 0.84). | CS maintained performance by leveraging its exploratory search mechanisms to mitigate noise effects. At higher noise levels, its exploratory search was somewhat less effective, leading to a slight performance reduction. However, it was still better at handling noise compared to local search-based algorithms, thanks to its ability to explore and adapt. |
| Bat Algorithm | Struggled with noise, showing significant drops in performance at 0.5 noise (accuracy of 0.82). | The Bat Algorithm's reliance on local search mechanisms made it susceptible to noise. With increasing noise levels, its optimization process failed to adapt, and it struggled to prioritize reliable clients effectively. As a result, its performance dropped significantly at the higher noise level. |
| Bee Colony | Moderate performance, with a slight decline under noise (accuracy of 0.81 at 0.5 noise). | Bee Colony's local search strategies did not allow it to handle adversarial noise effectively. As noise increased, the algorithm's inability to adapt to fluctuating data quality limited its performance, leading to a slight drop in accuracy and recall. |
| Ant Colony Optimization (ACO) | Poor performance, especially at 0.5 noise (accuracy of 0.79). | ACO's focus on local search and reliance on pheromone trails made it ineffective in noisy environments. The algorithm could not adjust to changes in data quality caused by noise, resulting in significant drops in performance at higher noise levels. |
| Fish Swarm | Significant accuracy drop, especially at 0.5 noise (accuracy of 0.8). | Fish Swarm's reliance on larger, more diverse client pools made it particularly vulnerable to noise. In noisy environments, its ability to select the best clients diminished, leading to poor performance as the algorithm struggled to adapt to data quality degradation. |
| Glow Worm | Performance significantly reduced, particularly at 0.5 noise (accuracy of 0.77). | Glow Worm's local optimization approach was highly vulnerable to noise. The algorithm's performance dropped substantially due to its reliance on localized search strategies, which were not robust enough to mitigate the impact of noise in the dataset. |
| Intelligent Water Droplet (IWD) | Similar to Glow Worm, IWD's performance dropped significantly under noise (accuracy of 0.76 at 0.5 noise). | IWD, like Glow Worm, utilized localized optimization, which was ineffective in the presence of adversarial noise. As a result, the performance of IWD was significantly impacted at both noise levels, with its ability to generalize under noisy conditions being severely compromised. |

## 5. Discussion

The results of this study highlight the exceptional performance of Grey Wolf Optimization (GWO) as the most effective Swarm Intelligence (SI) algorithm for client selection in Federated Learning (FL), particularly in the context of cybersecurity. GWO consistently outperformed other algorithms across all experimental scenarios, achieving the highest accuracy, recall, and F1-scores. Its dynamic exploration-exploitation balance and hierarchical decision-making framework allowed it to adapt effectively to varying client conditions, including noisy and heterogeneous environments. These characteristics make GWO a strong candidate for critical FL applications, such as intrusion detection in decentralized cybersecurity networks, where reliability and adaptability are paramount.

Particle Swarm Optimization (PSO) and Cuckoo Search (CS) also demonstrated robust performance in specific contexts, emerging as strong alternatives to GWO. PSO's iterative feedback mechanism enabled it to maintain competitive performance across dynamic configurations, making it well-suited for applications that require consistent adaptability, such as mobile healthcare systems or real-time industrial IoT networks. Similarly, CS proved effective in handling moderate noise and diverse data distributions, showing potential for resource-constrained environments, such as edge-based FL deployments. These algorithms, while not as versatile as GWO, are valuable in scenarios where specific constraints or conditions demand tailored optimization strategies.

In contrast, Glow Worm, Fish Swarm, and Intelligent Water Droplet (IWD) consistently underperformed across all scenarios. Their reliance on localized optimization strategies hindered their scalability and adaptability to dynamic and heterogeneous client conditions. Glow Worm, for example, demonstrated moderate effectiveness in static setups but struggled significantly in noisy or rapidly changing environments. Similarly, Fish Swarm and IWD lacked the ability to generalize effectively, rendering them less suitable for scenarios requiring robust and reliable optimization. These limitations underscore the importance of algorithm refinement for improving their utility in FL applications.

These findings emphasize the necessity of aligning algorithm selection with the specific requirements of FL applications. GWO's robustness makes it particularly suitable for cybersecurity systems, such as intrusion detection in decentralized networks, where it can manage heterogeneous data, noisy environments, and adversarial conditions while maintaining high accuracy and sensitivity. In contrast, PSO and CS are better suited for resource-constrained applications, such as industrial IoT and mobile healthcare systems, where efficiency and adaptability are critical. Although Glow Worm and Fish Swarm are currently less competitive, further refinement—such as enhancing exploration strategies or hybridizing with other algorithms—could improve their performance in lightweight deployments under stable conditions.

Future work could further optimize the performance of these algorithms through techniques such as hyperparameter tuning, hybrid approaches, and task-specific customizations. Hyperparameter tuning could systematically explore parameter configurations to enhance performance, particularly for underperforming algorithms like Fish Swarm and IWD. Hybrid approaches, such as combining GWO's hierarchical framework with PSO's feedback mechanism, could yield more robust solutions for complex FL scenarios. Additionally, task-specific customizations, tailored to address challenges in specific FL applications such as anomaly detection or resource optimization, could maximize their practical utility.

By linking these findings to actionable insights for practitioners, this study provides a foundation for applying and refining SI algorithms in diverse FL environments. The adaptability and scalability of these algorithms are particularly critical for FL deployment in cybersecurity, mobile healthcare, and IoT systems, where evolving requirements demand innovative and resilient solutions. These insights not only advance the current understanding of SI algorithms in FL but also offer a clear roadmap for future research and practical applications.

## 6. Limitations

While this study provides valuable insights into the performance of Swarm Intelligence (SI) algorithms for client selection in Federated Learning (FL), it is essential to acknowledge certain limitations. A key constraint lies in the use of synthetic data and simulated environments to evaluate algorithm performance. Although synthetic datasets allow for controlled experimentation and reproducibility, they may not fully capture the complexity and variability of real-world scenarios, such as dynamic data distributions, network conditions, and unforeseen adversarial attacks. Consequently, the results, while promising, may not translate seamlessly into practical deployments without further validation.

Another limitation is the simulation-based experimental setup, which assumes idealized conditions for communication and computational efficiency. In real-world FL systems, factors such as network latency, bandwidth constraints, and device heterogeneity could impact the algorithms' scalability and effectiveness. These practical considerations were not explicitly modeled in this study, potentially limiting the generalizability of the findings to large-scale, decentralized networks.

To address these limitations, future research should focus on validating the proposed algorithms in real-world FL deployments, particularly in domains such as IoT, healthcare, and finance. Incorporating real-world datasets and operational environments would provide a more comprehensive evaluation of the algorithms' robustness and adaptability. Additionally, exploring hybrid setups that combine synthetic simulations with real-world case studies could offer a balanced approach, ensuring both experimental control and practical relevance. Such efforts would strengthen the applicability of the findings and pave the way for scalable and resilient FL solutions in complex and dynamic systems.

## 7. Practical Implications and Future Work

The experiments demonstrated GWO's ability to excel in scenarios involving dynamic client participation, class imbalances, and noisy conditions, maintaining high performance regardless of client variability or adversarial challenges. PSO and Cuckoo Search ranked as strong alternatives, exhibiting balanced performance across metrics and computational efficiency, making them suitable for scalable deployments such as IoT systems in smart cities or industrial networks. While Bat and Bee Colony showed moderate effectiveness in controlled settings, algorithms like Fish Swarm, Glow Worm, and Intelligent Water Droplet struggled in scenarios requiring higher generalization, limiting their applicability to environments with stable and predictable client data.

These findings emphasize the practical value of SI algorithms in real-world FL applications across diverse industries. For example, GWO's robustness makes it an ideal choice for intrusion detection systems in decentralized cybersecurity networks, which often feature heterogeneous and noisy data. Similarly, PSO and Cuckoo Search's efficiency and scalability position them as strong candidates for industrial IoT deployments, such as smart factory networks or predictive maintenance systems. The versatility of these algorithms extends to critical applications in healthcare and finance, where reliable client selection can enhance predictive accuracy while maintaining data privacy and regulatory compliance. For practitioners, these results provide a clear framework for selecting algorithms tailored to specific FL scenarios. For example, GWO is recommended for critical infrastructure applications requiring high accuracy and adaptability, while PSO and Cuckoo Search are better suited for resource-constrained setups like mobile healthcare or large-scale IoT ecosystems.

The study also contributes novel insights to the field, representing the first systematic evaluation of underexplored SI algorithms, such as Fish Swarm and Glow Worm, in the context of FL client selection. This novelty highlights the potential for these algorithms to address specific challenges with further refinement, such as improving their generalization capabilities in dynamic and adversarial environments. By identifying strengths and limitations, the study provides actionable insights for practitioners to tailor algorithm selection based on unique FL requirements, such as optimizing for resource efficiency in constrained environments or maximizing detection accuracy in sensitive applications.

Future work should delve into hyperparameter fine-tuning to further enhance the performance of these algorithms. Techniques such as grid search or Bayesian Optimization can streamline the discovery of optimal configurations, ensuring peak performance for specific scenarios like dynamic client participation or non-IID data distributions. Additionally, expanding the scope of evaluation to include larger-scale networks with higher numbers of clients would offer insights into the scalability and robustness of these algorithms in national-level healthcare systems, global financial networks, or massive IoT ecosystems. Exploring hybrid approaches that integrate the strengths of multiple algorithms could also address limitations observed in weaker performers, creating more adaptive and resilient solutions. This study provides a solid foundation for advancing client selection in FL, paving the way for practical, robust, and scalable applications in decentralized systems across various domains.

## 8. Conclusion

This study systematically evaluated nine Swarm Intelligence (SI) algorithms for client selection in Federated Learning (FL) within cybersecurity contexts. Grey Wolf Optimization (GWO) emerged as the most robust performer, consistently achieving superior accuracy, recall, and F1-scores across diverse scenarios, including dynamic client participation, non-IID data distributions, and adversarial noise conditions. These findings underscore GWO's adaptability and scalability, making it an ideal candidate for real-world FL applications in cybersecurity, where reliability and resilience are critical. Particle Swarm Optimization (PSO) and Cuckoo Search (CS) also demonstrated strong performance, particularly in resource-constrained setups, highlighting their potential for applications in domains such as mobile healthcare and industrial IoT. In contrast, algorithms like Fish Swarm and Glow Worm, while less competitive, showed potential for improvement with further refinement.

The study's findings emphasize the importance of tailoring algorithm selection to specific FL challenges. For critical applications in cybersecurity, such as intrusion detection in decentralized networks, GWO's robustness offers a compelling solution for managing heterogeneous and noisy data. PSO and CS, with their adaptability and efficiency, are well-suited for scenarios requiring lightweight and scalable implementations. These results provide actionable insights for practitioners and contribute to advancing the practical application of SI algorithms in FL across various domains, including IoT, healthcare, and finance.

Beyond practical applications, this study contributes to the broader body of knowledge by systematically exploring the application of SI algorithms in FL, particularly in challenging cybersecurity contexts. By addressing gaps in the literature, such as the lack of optimization-driven approaches for client selection in FL, this work lays the groundwork for further exploration of SI-based techniques in decentralized learning systems. The insights gained from this research highlight the potential of SI algorithms to enhance the resilience, scalability, and adaptability of FL systems in complex and adversarial environments.

Future work should focus on several key areas to build upon these findings. Hyperparameter optimization could further enhance the performance of SI algorithms, especially underperforming ones like Fish Swarm and Glow Worm. Additionally, validating these algorithms with real-world datasets and operational deployments would strengthen their practical applicability. Exploring hybrid algorithm designs that combine the strengths of multiple SI methods, such as GWO's hierarchical decision-making with PSO's iterative feedback mechanisms, could yield innovative solutions for complex FL challenges. By pursuing these directions, the field of Federated Learning in cybersecurity is poised to evolve toward more effective, scalable, and resilient decentralized learning solutions.